\documentclass[11pt,a4paper,anonymous=False]{article}
\usepackage[hyperref]{emnlp-ijcnlp-2019}
\usepackage{times}
\usepackage{latexsym}

\usepackage{url}

\usepackage{booktabs} 
\usepackage{epstopdf}
\usepackage{amssymb}
\usepackage{amsmath}
\usepackage{graphicx}
\usepackage{multirow}
\usepackage{diagbox}

\usepackage{caption}
\usepackage{subfigure}

\usepackage{verbatim}
\usepackage{array}

\aclfinalcopy 

\title{Document Hashing with Mixture-Prior Generative Models}

\author{Wei Dong\textsuperscript{1}, \quad Qinliang Su\textsuperscript{1, 2}\thanks{\hspace{1mm} Corresponding author.}, \quad Dinghan Shen\textsuperscript{3}, \quad Changyou Chen\textsuperscript{4} \\
  \textsuperscript{1} School of Data and Computer Science, Sun Yat-sen University \\
  \textsuperscript{2} Guangdong Key Laboratory of Big Data Analysis and Processing, Guangzhou, China\\
    \textsuperscript{3} ECE Department, Duke University \quad
     \textsuperscript{4} CSE Department, SUNY at Buffalo \\
  {\tt dongw23@mail2.sysu.edu.cn, suqliang@mail.sysu.edu.cn} \\
  {\tt dinghan.shen@duke.edu, changyou@buffalo.edu } \\}

\date{}

\begin{document}
\maketitle
\begin{abstract}
  Hashing is promising for large-scale information retrieval tasks thanks to the efficiency of distance evaluation between binary codes. Generative hashing is often used to generate hashing codes in an unsupervised way. However, existing generative hashing methods only considered the use of simple priors, like Gaussian and Bernoulli priors, which limits these methods to further improve their performance. In this paper, two mixture-prior generative models are proposed, under the objective to produce high-quality hashing codes for documents. Specifically, a Gaussian mixture prior is first imposed onto the variational auto-encoder (VAE), followed by a separate step to cast the continuous latent representation of VAE into binary code. To avoid the performance loss caused by the separate casting, a model using a Bernoulli mixture prior is further developed, in which an end-to-end training is admitted by resorting to the straight-through (ST) discrete gradient estimator. Experimental results on several benchmark datasets demonstrate that the proposed methods, especially the one using Bernoulli mixture priors, consistently outperform existing ones by a substantial margin.
\end{abstract}

\section{Introduction}
Similarity search aims to find items that look most similar to the query one from a huge amount of data ~\cite{wang2018survey}, and are found in extensive applications like plagiarism analysis ~\cite{stein2007strategies}, collaborative filtering ~\cite{koren2008factorization,wang2016learning}, content-based multimedia retrieval ~\cite{lew2006content}, web services ~\cite{dong2004similarity} etc. Semantic hashing is an effective way to accelerate the searching process by representing every document with a compact binary code. In this way, one only needs to evaluate the hamming distance between binary codes, which is much cheaper than the Euclidean distance calculation in the original feature space.

Existing hashing methods can be roughly divided into data-independent and data-dependent categories. Data-independent methods employ random projections to construct hash functions without any consideration on data characteristics, like the locality sensitive hashing (LSH) algorithm \cite{datar2004locality}. On the contrary, data dependent hashing seeks to learn a hash function from the given training data in a supervised or an unsupervised way. In the supervised case, a deterministic function which maps the data to a binary representation is trained by using the provided supervised information (e.g. labels) ~\cite{liu2012supervised,shen2015supervised,liu2016deep}. However, the supervised information is often very difficult to obtain or is not available at all. Unsupervised hashing seeks to obtain binary representations by leveraging the inherent structure information in data, such as the spectral hashing ~\cite{weiss2009spectral}, graph hashing ~\cite{liu2011hashing}, iterative quantization ~\cite{gong2013iterative}, self-taught hashing ~\cite{zhang2010self} etc.

Generative models are often considered as the most natural way for unsupervised representation learning ~\cite{miao2016neural,bowman2015generating, yang2017improved}. Many efforts have been devoted to hashing by using generative models. In ~\cite{chaidaroon2017variational}, variational deep semantic hashing (VDSH) is proposed to solve the semantic hashing problem by using the variational autoencoder (VAE) ~\cite{kingma2013auto}. However, this model requires a two-stage training since a separate step is needed to cast the continuous representations in VAE into binary codes. Under the two-stage training strategy, the model is more prone to get stuck at poor performance ~\cite{xu2015convolutional,zhang2010self,wang2013semantic}. To address the issue, the neural architecture for generative semantic hashing (NASH) in ~\cite{shen2018nash} proposed to use a Bernoulli prior to replace the Gaussian prior in VDSH, and further use the straight-through (ST) method ~\cite{bengio2013estimating} to estimate the gradients of functions involving binary variables. It is shown that the end-to-end training brings a remarkable performance improvement over the two-stage training method in VDSH. Despite of superior performances, only the simplest priors are used in these models, i.e. Gaussian in VDSH and Bernoulli in NASH. However, it is widely known that priors play an important role on the performance of generative models  ~\cite{goyal2017nonparametric,chen2016variational,jiang2016variational}.

Motivated by this observation, in this paper, we propose to produce high-quality hashing codes by imposing appropriate mixture priors on generative models. Specifically, we first propose to model documents by a VAE with a Gaussian mixture prior. However, similar to the VDSH, the proposed method also requires a separate stage to cast the continuous representation into binary form, making it suffer from the same pains of two-stage training. Then we further propose to use a Bernoulli mixture as the prior, in hopes to yield binary representations directly. An end-to-end method is further developed to train the model, by resorting to the straight-through gradient estimator for neural networks involving binary random variables. Extensive experiments are conducted on benchmark datasets, which show substantial gains of the proposed mixture-prior methods over existing ones, especially the method with a Bernoulli mixture prior.

\begin{figure*}
	\centering
	\includegraphics[width=0.95\textwidth]{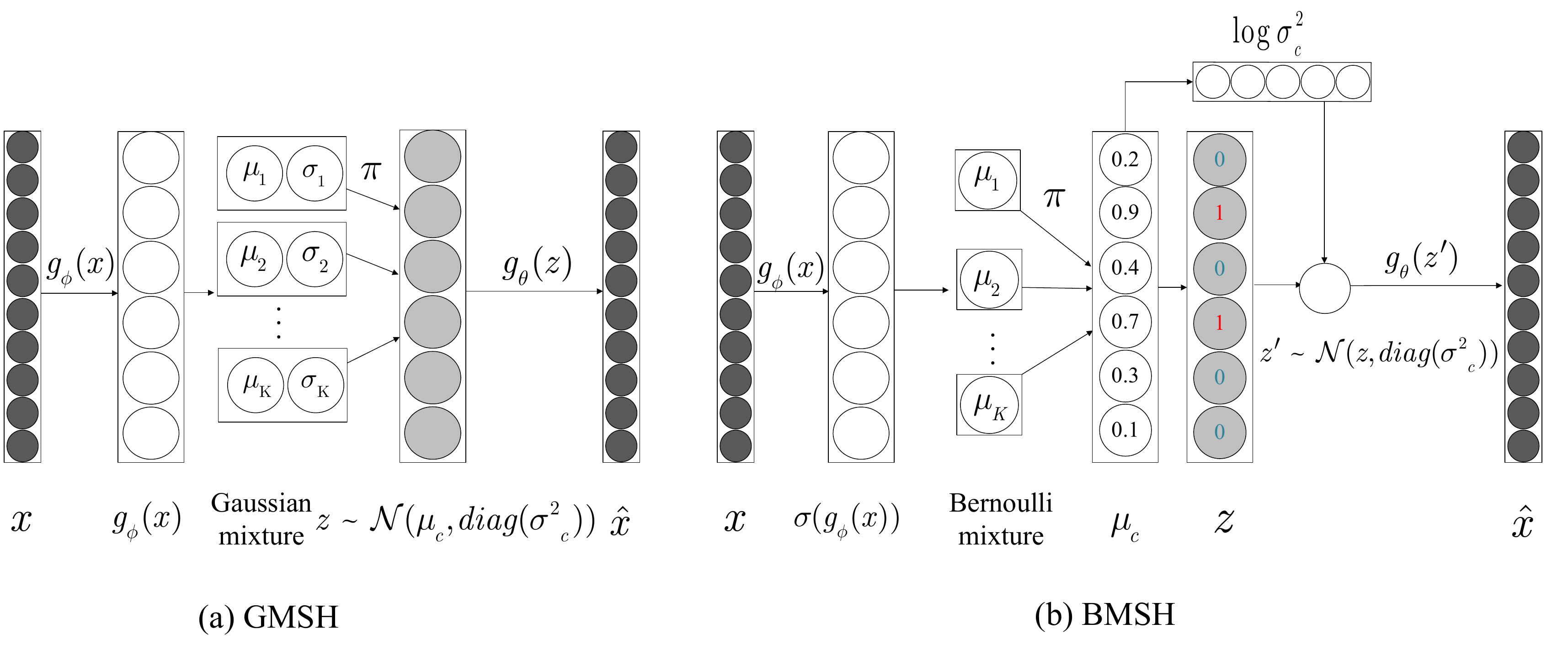}
    \vspace{-2.5mm}
	\caption{The architectures of the GMSH and BMSH. The data generative process of GMSH is done as follows: (1) Pick a component $c\in\{1,2,...,K\}$ from $Cat(\pi)$ with $\pi=[\pi_{1},\pi_{2},...,\pi_{K}]$; (2) Draw a sample $z$ from the picked Gaussian distribution $\mathcal{N} \left( \mu_{c}, diag(\sigma_{c}^{2})\right)$; (3) Use $g_{\theta}(z)$ to decode the sample $z$ into an observable $\hat x$. The process of generating data in BMSH can be described as follows: (1) Choose a component $c$ from $Cat(\pi)$; (2) Sample a latent vector from the chosen distribution $Bernoulli(\gamma_{c})$; (3) Inject data-dependent noise into $z$, and draw $z'$ from $\mathcal{N}(z, diag(\sigma_{c}^{2}))$; (4) Then use decoder $g_{\theta}(z')$ to reconstruct $\hat x$.}
	\label{fig:model}
\vspace{-5mm}
\end{figure*}

\section{Semantic Hashing by Imposing Mixture Priors}
In this section, we investigate how to obtain similarity-preserved hashing codes by imposing different mixture priors on variational encoder.
\subsection{Preliminaries on Generative Semantic Hashing}
Let $x \in \mathcal{Z}^{|V|}_{+}$ denote the bag-of-words representation of a document and $x_{i} \in \{0,1\}^{|V|}$ denote the one-hot vector representation of the $i$-{th} word of the document, where $|V|$ denotes the vocabulary size. VDSH in ~\cite{chaidaroon2017variational} proposed to model a document $\mathcal{D}$, which is defined by a sequence of one-hot word representations $\{x_i\}_{i=1}^{|\mathcal{D}|}$, with the joint PDF
\begin{equation} \label{pdf}
	p(\mathcal{D}, z)=p_{\theta}(\mathcal{D}|z)p(z),
\end{equation}
where the prior $p(z)$ is the standard Gaussian distribution ${\mathcal{N}}(0, I)$; the likelihood has the factorized form $p_{\theta}(\mathcal{D}|z)=\prod_{i=1}^{|\mathcal{D}|}p_{\theta}(x_i|z)$, and
\begin{equation} \label{decoder1}
	p_{\theta}(x_{i}|z) = \frac{\exp(z^{T}\mathnormal{E}x_{i}+b_{i})}{\sum_{j=1}^{|V|}\exp(z^{T}\mathnormal{E}x_{j}+b_{j})};
\end{equation}
$E\in {\mathbb{R}}^{m\times |V|}$ is a parameter matrix which connects latent representation $z$ to one-hot representation  $x_i$ of the $i$-th word, with $m$ being the dimension of $z$; $b_i$ is the bias term and $\theta = \{ E, b_{1}, ... , b_{|V|}\}$. It is known that generative models with better modeling capability often imply that the obtained latent representations are also more informative.

To increase the modeling ability of \eqref{pdf}, we may resort to more complex likelihood $p_{\theta}(\mathcal{D}|z)$, such as using deep neural networks to relate the latent $z$ to the observation $x_i$, instead of the simple softmax function in \eqref{decoder1}. However, as indicated in ~\cite{shen2018nash}, employing expressive nonlinear decoders likely destroy the distance-keeping property, which is essential to yield good hashing codes. In this paper, instead of employing a more complex decoder $p_{\theta}(\mathcal{D}|z)$, more expressive priors are leveraged to address this issue.
\subsection{Semantic Hashing by Imposing Gaussian Mixture Priors}\label{gm}
To begin with, we first replace the standard Gaussian prior $p(z)={\mathcal{N}}(0, I)$ in \eqref{pdf} by the following Gaussian mixture prior
\begin{equation}
	p(z) = \sum_{k=1}^K\pi_k \cdot {\mathcal{N}}\left(\mu_k, \text{diag}\left(\sigma_k^2\right)\right),
\end{equation}
where $K$ is the number of mixture components; $\pi_k$ is the probability of choosing the $k$-th component and $\sum_{k}^{K}\pi_{k} = 1$; $\mu_k\in {\mathbb{R}}^m$ and $\sigma_k^2\in {\mathbb{R}}_+^m$ are the mean and variance vectors of the Gaussian distribution of the $k$-th component; and $\text{diag}(\cdot)$ means diagonalizing the vector. 
For any sample $z \sim p(z)$, it can be equivalently generated by a two-stage procedure: 1) choosing a component $c\in \{1, 2, \cdots, K\}$ according to the categorical distribution $\text{Cat}(\pi)$  with $\pi = [\pi_1, \pi_2, \cdots, \pi_K]$; 2) drawing a sample from the distribution $\mathcal{N}\left(\mu_{c},\text{diag}\left(\sigma_c^2\right)\right)$. Thus, the document ${\mathcal{D}}$ is modelled as
\begin{equation}
	p({\mathcal{D}}, z, c) = p_{\theta}({\mathcal{D}}|z)p(z|c)p(c),
\end{equation}
where $p(z|c) = \mathcal{N}\left(\mu_{c},\text{diag}\left(\sigma_c^2\right)\right)$, $p(c) = \text{Cat}(\pi)$ and $p_{\theta}({\mathcal{D}}|z)$ is defined the same as \eqref{decoder1}.

To train the model, we seek to optimize the lower bound of the log-likelihood
\begin{equation}
\mathcal{L} = \mathnormal{E}_{q_{\phi}(z,c|x)}\left[\log\frac{p_{\theta}({\mathcal{D}}|z)p(z|c)p(c)}{q_{\phi}(z,c|x)}\right], \end{equation}
where $q_\phi(z, c|x)$ is the approximate posterior distribution of $p(z, c|x)$ parameterized by $\phi$; here $x$ could be any representation of the documents, like the bag-of-words, TFIDF etc. For the sake of tractability, $q_{\phi}(z,c|x)$ is further assumed to maintain a factorized form, i.e.,  $q_{\phi}(z,c|x) = q_\phi(z|x)q_\phi(c|x)$. Substituting it into the lower bound gives
\begin{align} \label{GM_elbo}
	{\mathcal{L}} =& {\mathbb{E}}_{q_\phi(z|x)}\left[\log p_{\theta}({\mathcal{D}}|z) \right] - KL\left(q_\phi(c|x) || p(c)\right) \nonumber \\
	&- {\mathbb{E}}_{q_\phi(c|x)} \left[ KL\left(q_\phi(z|x) || p(z|c) \right) \right].
\end{align}
For simplicity, we assume that $q_\phi(z|x)$ and $q_\phi(c|x)$ take the forms of Gaussian and categorical distributions, respectively, and the distribution parameters are defined as the outputs of neural networks. The entire model, including the generative and inference arms, is illustrated in Figure \ref{fig:model}(a). Using the properties of Gaussian and categorical distributions, the last two terms in \eqref{GM_elbo} can be expressed in a closed form. Combining with the reparameterization trick in stochastic gradient variational bayes (SGVB) estimator \cite{kingma2013auto}, the lower bound ${\mathcal{L}}$ can be optimized w.r.t. model parameters $\{\theta, \pi, \mu_k, \sigma_k, \phi\}$ by error backpropagation and SGD algorithms directly.

Given a document $x$, its hashing code can be obtained through two steps: 1) mapping $x$ to its latent representation by $z=\mu_{\phi}(x)$, where the $\mu_{\phi}(x)$ is the encoder mean $\mu_{\phi}(\cdot)$; 2) thresholding $z$ into binary form. As suggested in ~\cite{wang2013semantic,chaidaroon2018deep,chaidaroon2017variational} that when hashing a batch of documents, we can use the median value of the elements in $z$ as the critical value, and threshold each element of $z$ into $0$ and $1$ by comparing it to this critical value. For presentation conveniences, the proposed semantic hashing model with a Gaussian mixture priors is referred as GMSH.

\subsection{Semantic Hashing by Imposing Bernoulli Mixture Priors}
To avoid the separate casting step used in GMSH, inspired by NASH~\cite{shen2018nash}, we further propose a {\it{S}}emantic {\it{H}}ashing model with a {\it{B}}ernoulli {\it{M}}ixture prior (BMSH). Specifically, we replace the Gaussian mixture prior in GMSH with the following Bernoulli mixture prior
\begin{equation}
	p(z) = \sum_{k=1}^K\pi_k \cdot \text{Bernoulli}\left(\gamma_k\right),
\end{equation}
where $\gamma_k \in [0, 1]^m$ represents the probabilities of $z$ being 1.
Effectively, the Bernoulli mixture prior, in addition to generating discrete samples, plays a similar role as Gaussian mixture prior, which make the samples drawn from different components have different patterns. The samples from the Bernoulli mixture can be generated by first choosing a component $c\in \{1, 2, \cdots, K\}$ from $\text{Cat}(\pi)$  and then drawing a sample from the chosen  distribution $\text{Bernoulli}(\gamma_c)$.
The entire model can be described as
$
p({\mathcal{D}},z,c) = p_{\theta}({\mathcal{D}}|z) p(z|c)p(c),
$
where $ p_{\theta}({\mathcal{D}}|z)$ is defined the same as \eqref{decoder1}, and $p(c)  = \text{Cat}(\pi)$ and $p(z|c) = \text{Bernoulli}(\gamma_{c}).$

Similar to GMSH, the model can be trained by maximizing the variational lower bound, which maintains the same form as \eqref{GM_elbo}. Different from GMSH, in which $q_\phi(z|x)$ and $p(z|c)$ are both in a Gaussian form, here $p(z|c)$ is a Bernoulli distribution by definition, and thus $q_\phi(z|x)$ is assumed to be the Bernoulli form as well, with the probability of the $i$-th element $z_i$ taking 1 defined as
\begin{equation}
	q_\phi(z_i=1|x) \triangleq \sigma\left(g_\phi^i(x)\right)
\end{equation}
for $i=1, 2, \cdots, m$. Here $g_\phi^i(\cdot)$ indicates the $i$-th output of a neural network parameterized by $\phi$.
Similarly, we also define the posterior regarding which component to choose as
\begin{equation}
	q_\phi(c = k |x) = \frac{\exp\left(h_\phi^k(x)\right)}{\sum_{i=1}^K\exp\left(h_\phi^i(x)\right)},
\end{equation}
where $h_\phi^k(x)$ is the $k$-th output of a neural network parameterized by $\phi$.
With denotation $\alpha_i = q_\phi(z_i=1|x)$ and $\beta_k = q_\phi(c=k|x)$, the last two terms in \eqref{GM_elbo} can be expressed in close-form as
\begin{equation*}
	KL\left(q_\phi(c|x)||p(c)\right) = \sum_{c=1}^{K}\beta_{c}\log\frac{\beta_{c}}{\pi},
\end{equation*}
\vspace{-4mm}
\begin{equation*}
\begin{split}
&\mathbb{E}_{q_\phi(c|x)}\left[ KL\left(q_\phi(z|x) || p(z|c) \right) \right]\\
&=\!\sum_{c=1}^{K}\beta_{c}\!\sum_{i=1}^{m}\! \left(\! \alpha_{i} \log \!\! \frac{ \alpha_{i} }{ \gamma_{c}^{i}}+\! (1 \!- \alpha_{i}) \log \! \frac{1 \!- \alpha_{i}}{1 \!- \gamma_{c}^{i} }\right),
\end{split}
\end{equation*}
where $\gamma_{c}^{i}$ denotes the $i$-th element of $\gamma_{c}$.

Due to the Bernoulli assumption for the posterior $q_\phi(z|x)$, the commonly used reparameterization trick for Gaussian distribution cannot be used to directly estimate the first term ${\mathbb{E}}_{q_\phi(z|x)}\left[\log p_{\theta}({\mathcal{D}}|z) \right] $ in \eqref{GM_elbo}. Fortunately, inspired by the straight-through gradient estimator in \cite{bengio2013estimating}, we can parameterize the $i$-th element of binary sample $z$ from $q_\phi(z|x)$ as
\begin{equation}\label{x_to_z_BM}
	z_i = 0.5\times \left(sign \left(\sigma( g_\phi^i(x) ) - \xi_i \right) + 1\right),
\end{equation}
where $sign(\cdot)$ the is the sign function, which is equal to 1 for nonnegative inputs and -1 otherwise; and $\xi_i \sim \text{Uniform}(0, 1)$ is a uniformly random sample between 0 and 1.

The reparameterization method used above can guarantee generating binary samples. However, backpropagation cannot be used to optimize the lower bound ${\mathcal{L}}$ since the gradient of $sign(\cdot)$ w.r.t. its input is zero almost everywhere. To address this problem, the straight-through(ST) estimator  ~\cite{bengio2013estimating} is employed to estimate the gradient for the binary random variables, where the derivative of $z_i$ w.r.t $\phi$ is simply approximated by $0.5\times\frac{\partial \sigma(g_\phi^i(x))}{\partial \phi}$. Thus, the gradients can then be backpropagated through discrete variables. Similar to NASH ~\cite{shen2018nash}, data-dependent noises are also injected into the latent variables when reconstructing the document $x$ so as to obtain more robust binary representations. The entire model of BMSH, including generative and inference parts, is illustrated in Figure \ref{fig:model}(b).

To understand how the mixture-prior model works differently from the simple prior model, we examine the main difference term ${\mathbb{E}}_{q_\phi(c|x)} \left[ KL\left(q_\phi(z|x) || p(z|c) \right) \right]$ in \eqref{GM_elbo}, where $q_{\phi}(c|x)$ is the approximate posterior probability that indicates the document $x$ is generated by the $c$-th component distribution with $c\in \{1, 2, \cdots, K\}$. In the mixture-prior model, the approximate posterior $q_\phi(z|x)$ is compared to all mixture components $p(z|c)={\mathcal{N}}\left(\mu_c, \text{diag}(\sigma^2_c)\right)$. The term ${\mathbb{E}}_{q_\phi(c|x)} \left[ KL\left(q_\phi(z|x) || p(z|c) \right) \right]$ can be understood as the average of all these KL-divergences weighted by the probabilities $q_{\phi}(c|x)$. Thus, comparing to the simple-prior model, the mixture-prior model is endowed with more flexibilities, allowing the documents to be regularized by different mixture components according to their context.

\subsection{Extensions to Supervised Hashing}
When label information is available, it can be leveraged to yield more effective hashing codes since labels provide extra information about the similarities of documents. Specifically, a mapping from the latent representation $z$ to the corresponding label $y$ is learned for each document. The mapping encourages latent representations of documents with the same label to be close in the latent space, while those with different labels to be distant. A classifier built from a two-layer MLP is employed to parameterize this mapping, with its cross-entropy loss denoted by ${\mathcal{L}}_{dis}(z, y)$. Taking the supervised objective into account, the total loss is defined as
\begin{equation}\label{eqsupervised}
	{\mathcal{L}}_{total} = -{\mathcal{L}} +\alpha {\mathcal{L}}_{dis}(z, y),
\end{equation}
where ${\mathcal{L}} $ is the lower bound arising in GMSH or BMSH model; $\alpha$ controls the relative weight of the two losses. By examining the total loss ${\mathcal{L}}_{total}$, it can be seen that minimizing the loss encourages the model to learn a representation $z$ that accounts for not only the unsupervised content similarities of documents, but also the supervised similarities from the extra label information.
\begin{table*}
\vspace{-3mm}
\resizebox{\textwidth}{23mm}{
  \begin{tabular}{|c|cccc|cccc|cccc|}
    \toprule
    Datasets&\multicolumn{4}{c|}{TMC}&\multicolumn{4}{c|}{20Newsgroups}&\multicolumn{4}{c|}{Reuters}\\
    \midrule
    Method&16bit&32bit&64bit&128bit&16bit&32bit&64bit&128bit&16bit&32bit&64bit&128bit\\
    \midrule
    LSH &0.4393&0.4514&0.4553&0.4773&0.0597&0.0666&0.0770&0.0949&0.3215&0.3862&0.4667&0.5194\\
    S-RBM&0.5108&0.5166&0.5190&0.5137&0.0604&0.0533&0.0623&0.0642&0.5740&0.6154&0.6177&0.6452\\
    SpH &0.6055&0.6281&0.6143&0.5891&0.3200&0.3709&0.3196&0.2716&0.6340&0.6513&0.6290&0.6045\\
    STH&0.3947&0.4105&0.4181&0.4123&0.5237&0.5860&0.5806&0.5433&0.7351&0.7554&0.7350&0.6986\\
    VDSH&0.6853&0.7108&0.4410&0.5847&0.3904&0.4327&0.1731&0.0522&0.7165&0.7753&0.7456&0.7318\\
    NASH&0.6573&0.6921&0.6548&0.5998&0.5108&0.5671&0.5071&0.4664&0.7624&0.7993&0.7812&0.7559\\
    \hline
    GMSH&0.6736&0.7024&0.7086&0.7237&0.4855&0.5381&0.5869&0.5583&0.7672&0.8183&0.8212&0.7846\\
    BMSH&\textbf{0.7062}&\textbf{0.7481}&\textbf{0.7519}&\textbf{0.7450}&\textbf{0.5812}&\textbf{0.6100}&\textbf{0.6008}&\textbf{0.5802}&\textbf{0.7954}&\textbf{0.8286}&\textbf{0.8226}&\textbf{0.7941}\\
  \bottomrule
  \end{tabular}}
\caption{The precisions of the top 100 retrieved documents on three datasets with different numbers of hashing bits in unsupervised hashing.}
\vspace{-3mm}
\label{tab:results}
\end{table*}

\begin{figure*}[htbp]
\centering
\begin{minipage}{5cm}
\centering
\includegraphics[scale=0.33]{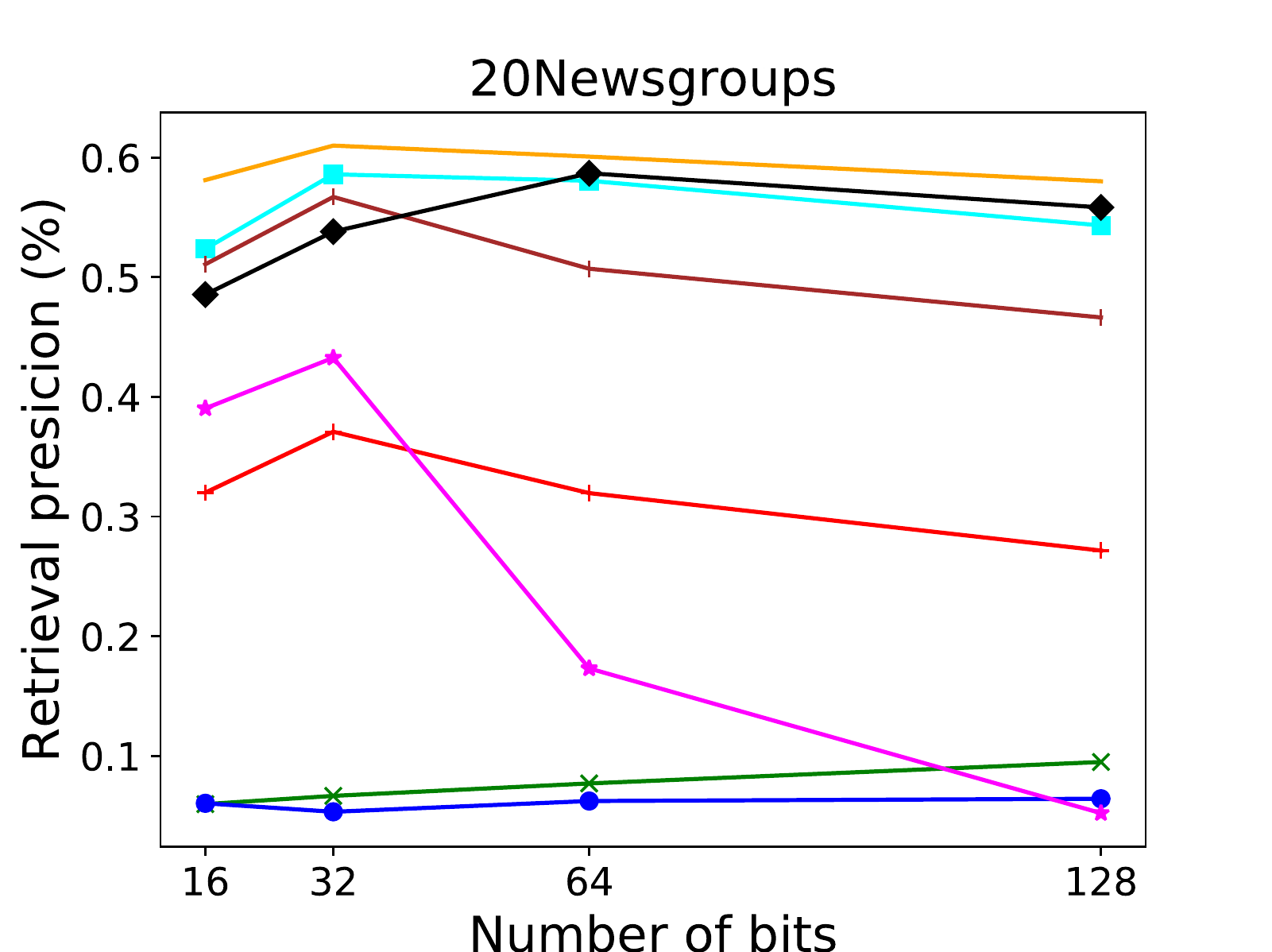}
\end{minipage}
\begin{minipage}{5cm}
\centering
\includegraphics[scale=0.33]{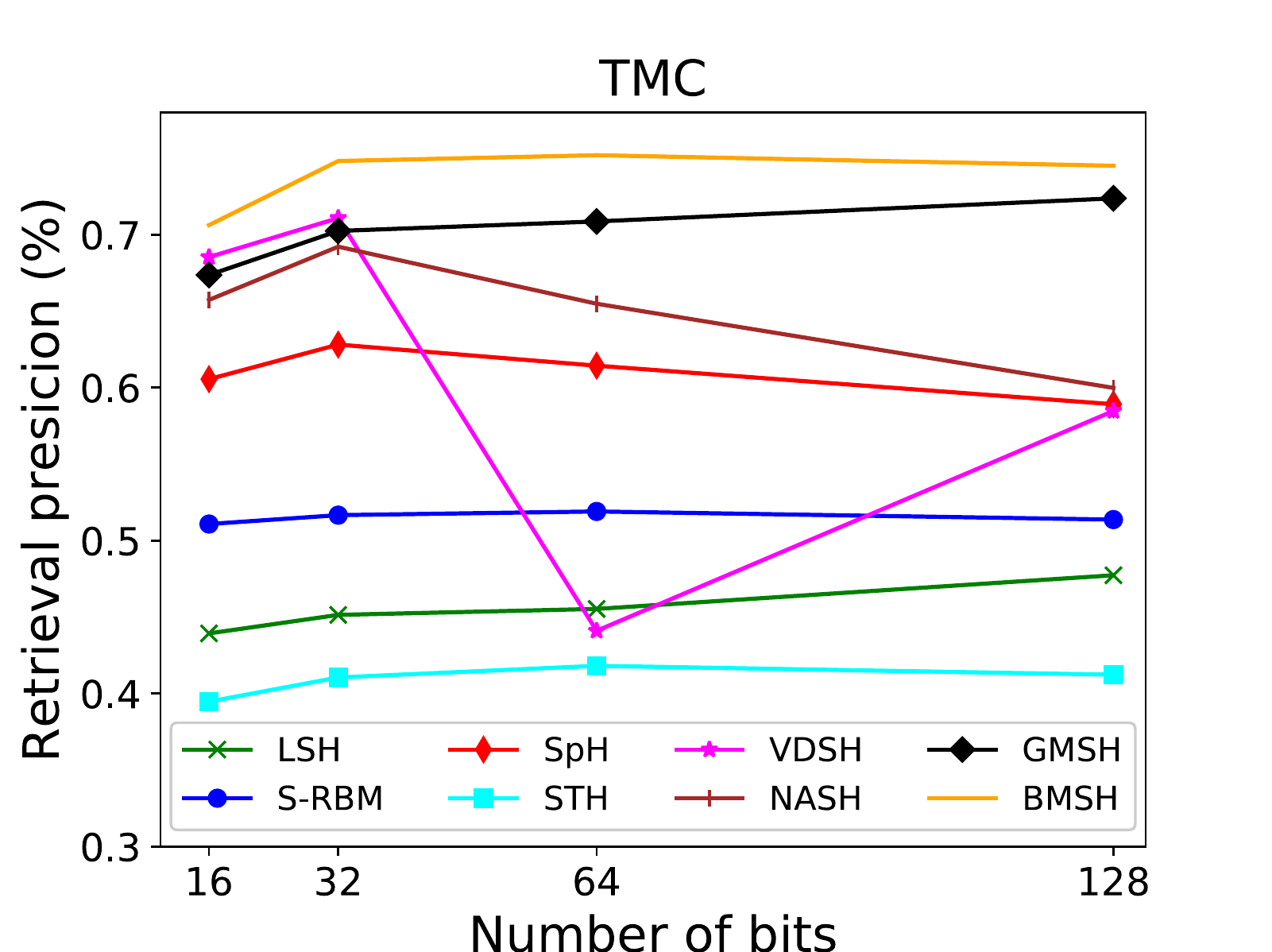}
\end{minipage}
\begin{minipage}{5cm}
\centering
\includegraphics[scale=0.33]{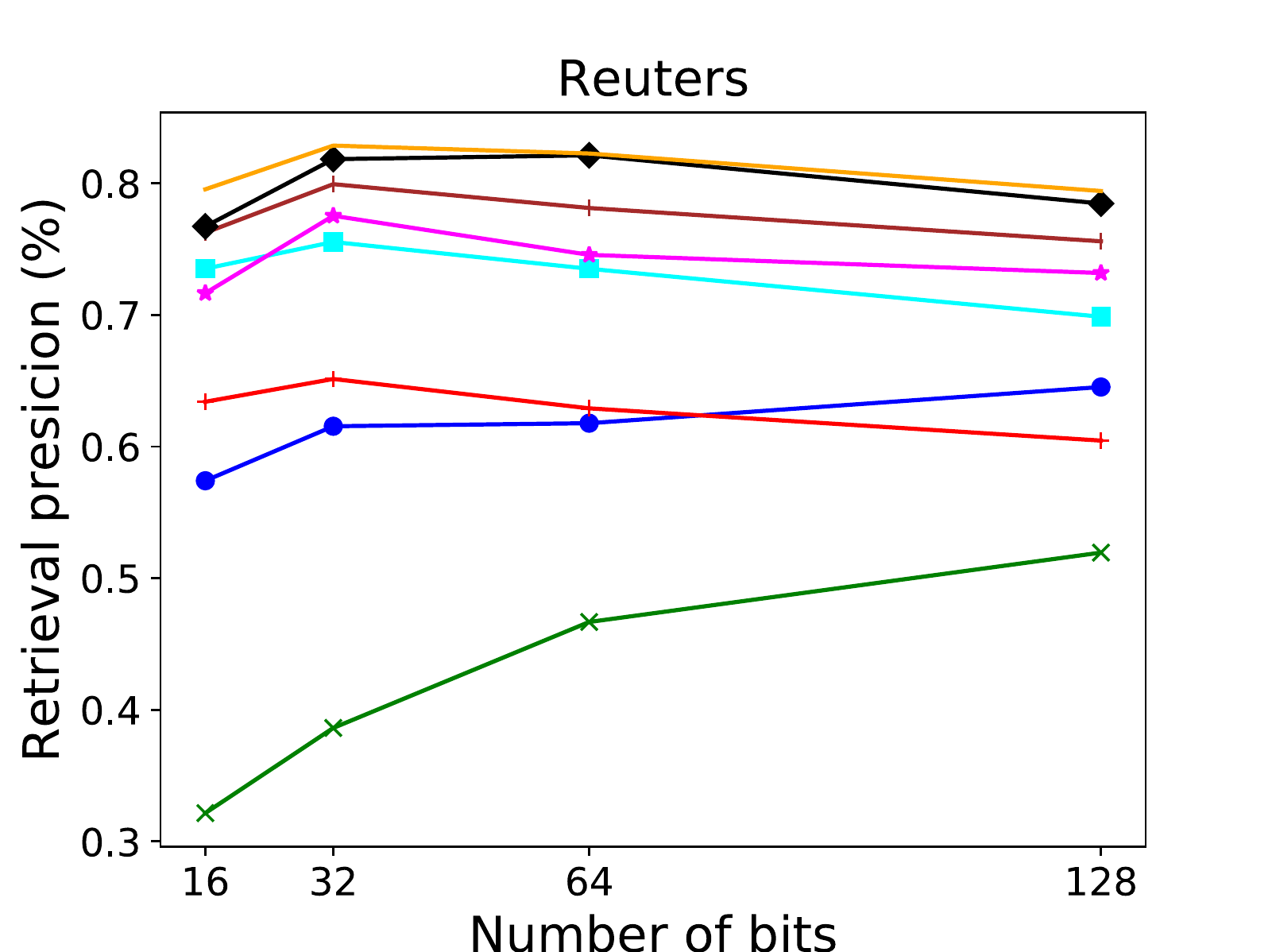}
\end{minipage}
\caption{The performance of unsupervised hashing models on three datasets with various numbers of hashing bits.}
\label{drop}
\vspace{-5mm}
\end{figure*}

\section{Related Work}
Existing hashing methods can be categorized into data independent and data dependent methods. A typical example of data independent hashing is the local-sensitive hashing (LSH) ~\cite{datar2004locality}. However, such method usually requires long hashing codes to achieve satisfactory performance. To yield more effective hashing codes, more and more researches focus on data dependent hashing methods, which include unsupervised and supervised methods. Unsupervised hashing methods only use unlabeled data to learn hash functions. For example, spectral hashing (SpH) ~\cite{weiss2009spectral} learns the hash function by imposing balanced and uncorrelated constraints on the learned codes. Iterative quantization (ITQ) ~\cite{gong2013iterative} generates the hashing codes by simultaneously maximizing the variance of each binary bit and minimizing the quantization error. In ~\cite{zhang2010self}, the authors proposed to decompose the learning procedure into two steps: first learning hashing codes for documents via unsupervised learning, then using $\ell$ binary classifiers to predict the $\ell$-bit hashing codes. Since the labels provide useful guidance in learning effective hash functions, supervised hashing methods are proposed to leverage the label information. For instance, binary reconstruction embedding (BRE) ~\cite{kulis2009learning} learns the hash function by minimizing the reconstruction error between the original distances and the hamming distances of the corresponding hashing codes. Supervised hashing with kernels (KSH) ~\cite{liu2012supervised} is a kernel-based method, which utilizes the pairwise information between samples to generate hashing codes by minimizing the hamming distances on similar pairs and maximizing those on dissimilar pairs.

Recently, VDSH ~\cite{chaidaroon2017variational} proposed to use a VAE to learn the latent representations of documents and then use a separate stage to cast the continuous representations into binary codes. While fairly successful, this generative hashing model requires a two-stage training. NASH ~\cite{shen2018nash} proposed to substitute the Gaussian prior in VDSH with a Bernoulli prior to tackle this problem, by using a straight-through estimator ~\cite{bengio2013estimating} to estimate the gradient of neural network involving the binary variables. This model can be trained in an end-to-end manner. Our models differ from VDSH and NASH in that mixture priors are employed to yield better hashing codes, whereas only the simplest priors are used in both VDSH and NASH.

\section{Experiments}
\subsection{Experimental Setups}
{\textbf{Datasets}}
Three public benchmark datasets are used in our experiments. $i)\ Reuters21578$: A dataset consisting of 10788 news documents from 90 different categories; $ii)\ 20Newsgroups$: A collection of 18828 newsgroup posts that are divided into 20 different newsgroups; $iii)\ TMC$: A dataset containing the air traffic reports provided by NASA, which includes 21519 training documents with 22 labels.

{\textbf{Training Details}}
We experiment with the four models proposed in this paper, i.e., GMSH and BMSH for unsupervised hashing, and GMSH-S and BMSH-S for supervised hashing. The same network architectures as VDSH and NASH are used in our experiments to admit a  fair comparison. Specifically, a two-layer feed-forward neural network with 500 hidden units and ReLU activation function is employed as the encoder and the extra classifier in the supervised case, while the decoder is the same as that stated in \eqref{decoder1}. Similar to VDSH and NASH \cite{chaidaroon2017variational,shen2018nash}, the TFIDF feature of a document is used as the input to the encoder. The Adam optimizer \cite{kingma2014adam} is used in the training of our models, and its learning rate is set to be $1 \times 10^{-3}$, with a decay rate of 0.96 for every 10000 iterations. The component number $K$ and the parameter $\alpha$ in \eqref{eqsupervised} are determined based on the validation set.

{\textbf{Baselines}}
For unsupervised semantic hashing, we compare the proposed GMSH and BMSH with the following models: locality sensitive hashing (LSH), stack restricted boltzmann machines (S-RBM), spectral hashing (SpH), self-taught hashing (STH), variational deep semantic hashing (VDSH) and neural architecture for semantic hashing(NASH). For supervised semantic hashing, we also compare GMSH-S and BMSH-S with the following baselines: supervised hashing with kernels (KSH) ~\cite{liu2012supervised}, semantic hashing using tags and topic modeling (SHTTM) ~\cite{wang2013semantic}, supervised VDSH and supervised NASH.

\begin{table*}
\resizebox{\textwidth}{18mm}{
  \begin{tabular}{|c|cccc|cccc|cccc|}
    \toprule
    Datasets&\multicolumn{4}{c|}{TMC}&\multicolumn{4}{c|}{20Newsgroups}&\multicolumn{4}{c|}{Reuters}\\
    \midrule
    Method&16bit&32bit&64bit&128bit&16bit&32bit&64bit&128bit&16bit&32bit&64bit&128bit\\
    \midrule
    KSH&0.6842&0.7047&0.7175&0.7243&0.5559&0.6103&0.6488&0.6638&0.8376&0.8480&0.8537&0.8620\\
    SHTTM&0.6571&0.6485&0.6893&0.6474&0.3235&0.2357&0.1411&0.1299&0.8520&0.8323&0.8271&0.8150\\
    VDSH-S&0.7887&0.7883&0.7967&0.8018&0.6791&0.7564&0.6850&0.6916&0.9121&0.9337&0.9407&0.9299\\
    NASH-DN-S&0.7946&0.7987&0.8014&0.8139&0.6973&0.8069&0.8213&0.7840&0.9327&0.9380&0.9427&0.9336\\
    \hline
    GMSH-S&0.7806&0.7929&0.8103&0.8144&0.6972&0.7426&0.7574&0.7690&0.9144&0.9175&0.9414&0.9522\\
    BMSH-S&\textbf{0.8051}&\textbf{0.8247}&\textbf{0.8340}&\textbf{0.8310}&\textbf{0.7316}&\textbf{0.8144}&\textbf{0.8216}&\textbf{0.8183}&\textbf{0.9350}&\textbf{0.9640}&\textbf{0.9633}&\textbf{0.9590}\\
  \bottomrule
  \end{tabular}}
\caption{The performances of different supervised hashing models on three datasets under different lengths of hashing codes.}
\vspace{-3mm}
\label{tab:su_results}
\end{table*}

\begin{figure*}[]
\centering
\begin{minipage}{5cm}
\centering
\includegraphics[scale=0.330]{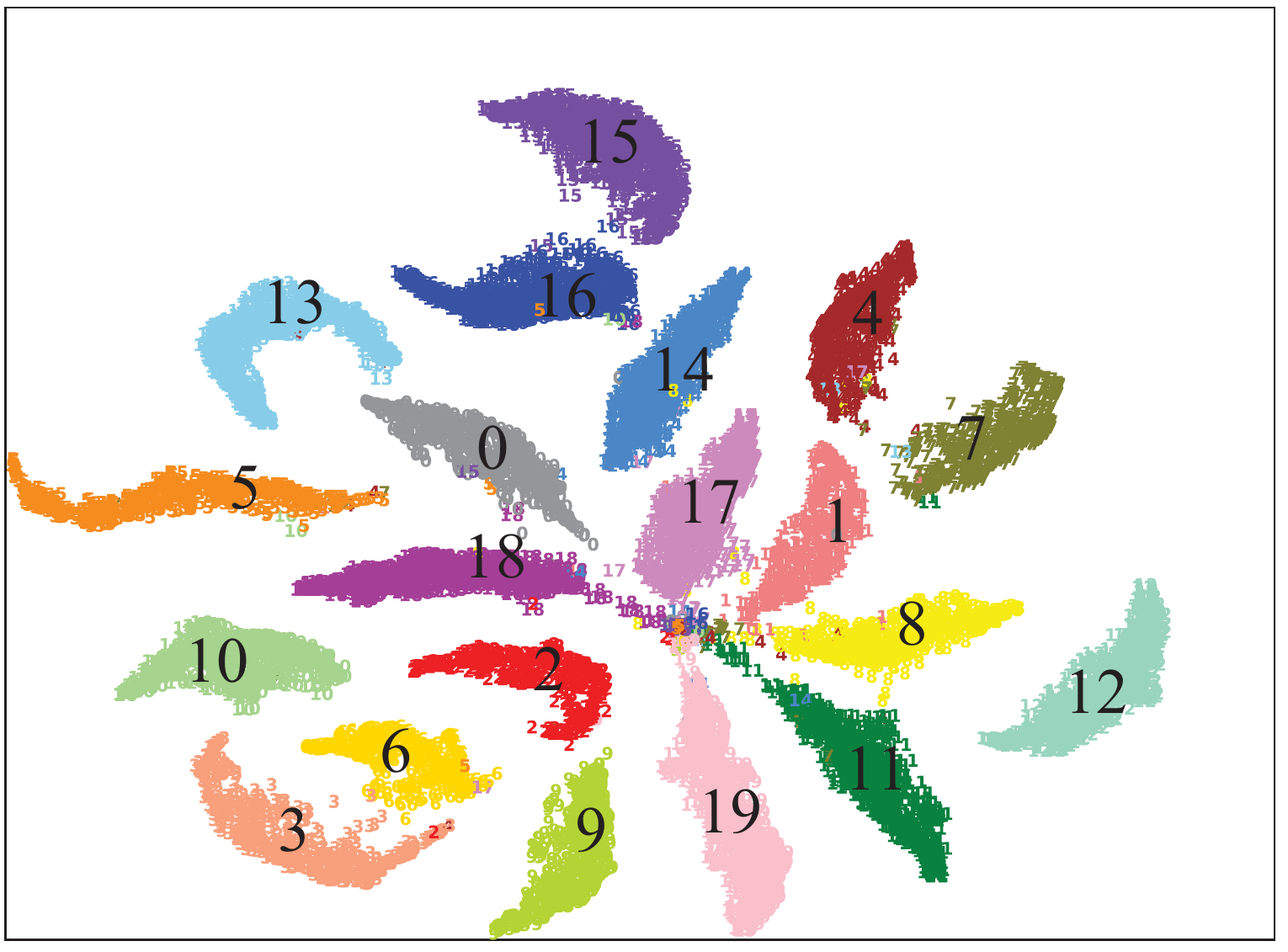}
\caption*{(a) VDSH-S}
\end{minipage}
\begin{minipage}{5cm}
\centering
\includegraphics[scale=0.33]{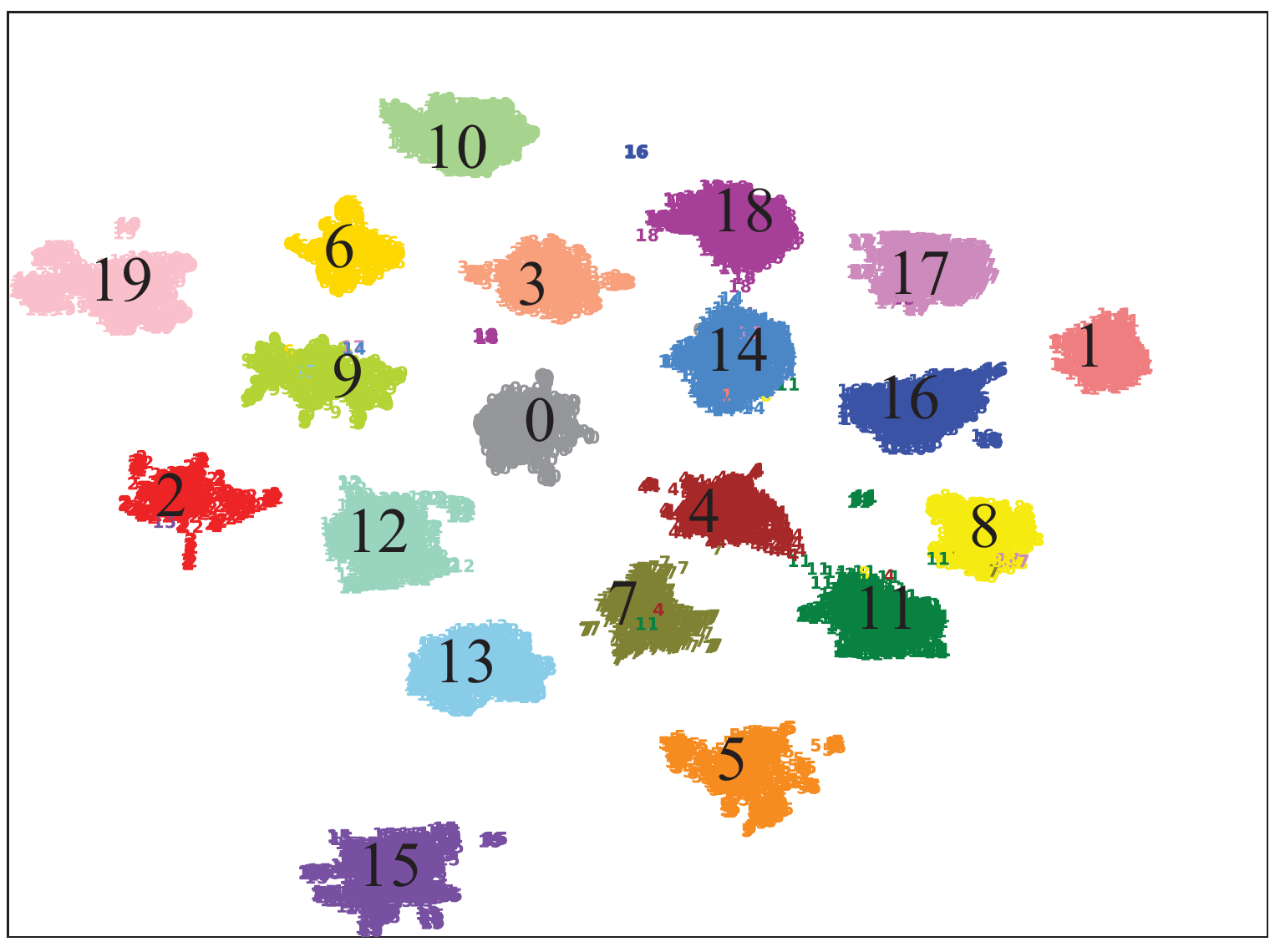}
\caption*{(b) GMSH-S}
\end{minipage}
\begin{minipage}{5cm}
\centering
\includegraphics[scale=0.33]{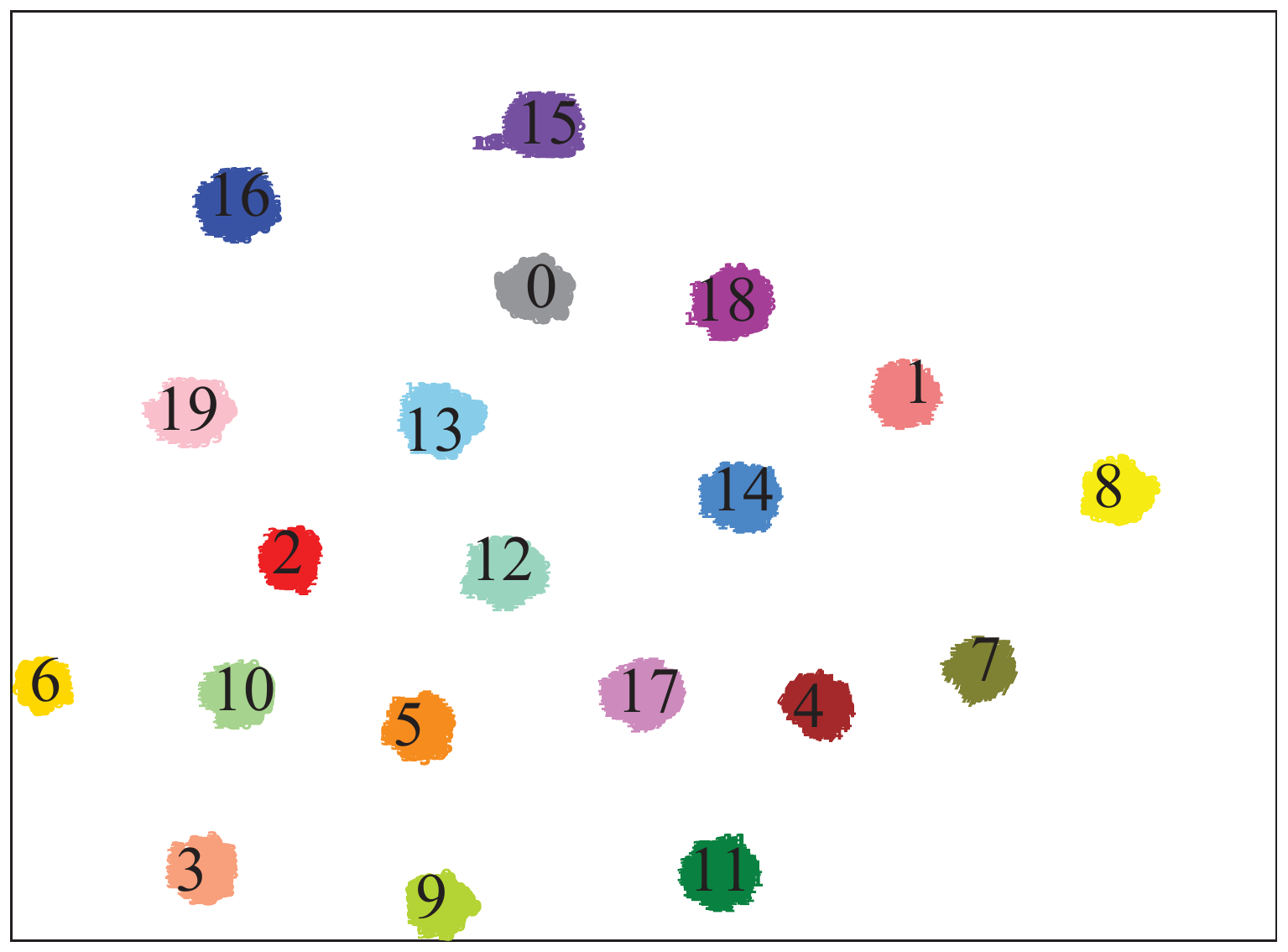}
\caption*{(c) BMSH-S}
\end{minipage}
\caption{Visualization of the 32-dimensional document latent semantic embeddings learned by VDSH-S, GMSH-S and MBSH-S on 20Newsgroups dataset. Each data point in the figure denotes a document, with each color representing one category. The number shown with the color is the ground-true category ID.}
\vspace{-4mm}
\label{fig:embedding}
\end{figure*}
{\textbf{Evaluation Metrics}}
For every document from the testing set, we retrieve similar documents from the training set based on the hamming distance between their hashing codes. For each query, 100 closest documents are retrieved, among which the documents sharing the same label as the query are deemed as the relevant results. The ratio between the number of relevant ones and the total number, which is 100, is calculated as the similarity search precision. The averaged value over all testing documents is then reported. The retrieval precisions under the cases of 16 bits, 32 bits, 64 bits, 128 bits hashing codes are evaluated, respectively.

\subsection{Performance Evaluation of Unsupervised Semantic Hashing}
Table \ref{tab:results} shows the performance of the proposed and baseline models on three datasets under the unsupervised setting, with the number of hashing bits ranging from 16 to 128. From the experimental results, it can be seen that GMSH outperforms previous models under all considered scenarios on both TMC and Reuters. It also achieves better performance on 20Newsgroups when the length of hashing codes is large, e.g. 64 or 128. Comparing to VDSH using the simple Gaussian prior, the proposed GMSH using a Gaussian mixture prior exhibits better retrieval performance overall. This strongly demonstrates the benefits of using mixture priors on the task of semantic hashing. One possible explanation is that the mixture prior enables the documents from different categories to be regularized by different distributions, guiding the model to learn more distinguishable representations for documents from different categories. It can be further observed that among all methods, BMSH achieves the best performance under different datasets and hashing codes length consistently. This may be attributed to the imposed Bernoulli mixture prior, which offers both the advantages of producing more distinguishable codes with a mixture prior and end-to-end training enabled by a Bernoulli prior. BMSH integrates the merits of NASH and GMSH, and thus is more suitable for the hashing task.

\begin{figure}
\centering
\includegraphics[width=0.4\textwidth]{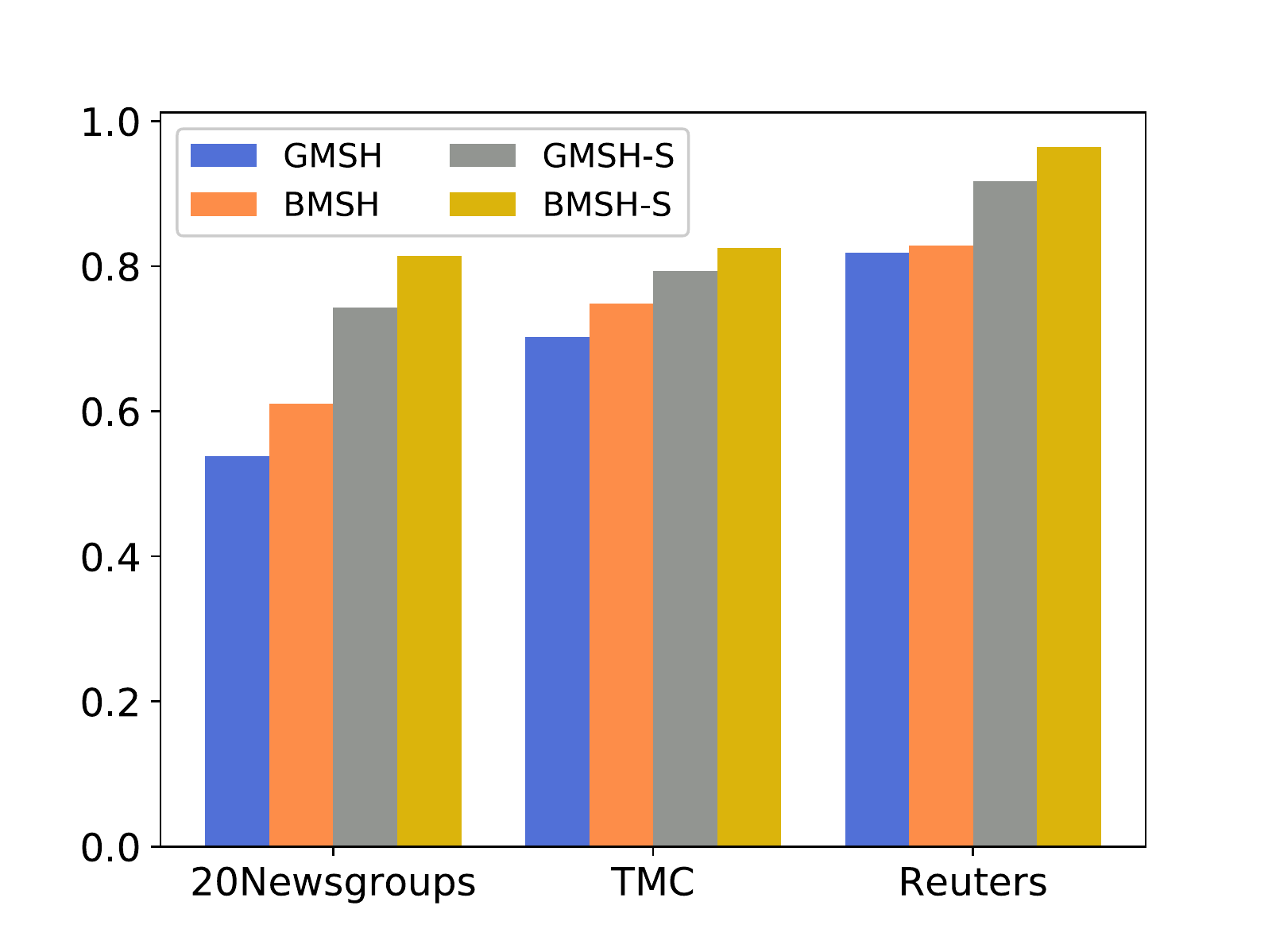}
\caption{The retrieval precisions of GMSH and BMSH on three datasets in both unsupervised and supervised scenarios.}
\label{fig:supervised_ours}
\vspace{-5mm}
\end{figure}

Figure \ref{drop} shows how retrieval precisions vary with the number of hashing bits on the three datasets. It can be observed that as the number increases from 32 to 128, the retrieval precisions of most previous models tend to decrease. This phenomenon is especially obvious for VDSH, in which the precisions on all three datasets drop by a significant margin. This interesting phenomenon has been reported in previous works  \cite{shen2018nash,chaidaroon2017variational,wang2013semantic,liu2012supervised}, and the reason could be overfitting since the model with long hashing codes is more likely to overfitting \cite{chaidaroon2017variational,shen2018nash}. However, it can be seen that our model is more robust to the number of hashing bits. When the number is increased to 64 or 128, the performance of our models is kept almost unchanged. This may be also attributed to the mixture priors imposed in our models, which can regularize the models more effectively.
\begin{table*}
  \small
  \centering
  \begin{tabular}{|c|cc|cc|cc|}
    \hline
    \multirow{2}*{\diagbox{$K$}{D}}&\multicolumn{2}{c|}{20Newsgroups}&\multicolumn{2}{c|}{TMC}&\multicolumn{2}{c|}{Reuters}\\
    \cline{2-7}
    ~&GMSH&BMSH&GMSH&BMSH&GMSH&BMSH\\
    \hline
    5&0.4708&0.5977&0.6886&0.7492&0.7888&0.8152\\
    10&0.4778&0.6007&0.6862&0.7479&0.8039&0.8226\\
    20&0.5381&0.6100&0.6883&\textbf{0.7495}&0.8182&0.\textbf{8286}\\
    40&0.5197&0.6015&0.7024&0.7481&0.8169&0.8258\\
    80&0.5188&0.6012&\textbf{0.7033}&0.7467&0.8087&0.8253\\
    GT&\textbf{0.5381}&\textbf{0.6100}&0.6960&0.7443&\textbf{0.8183}&0.8279\\
  \hline
  \end{tabular}
\caption{Precisions of top 100 retrieved documentswith different numberer of clusters, $K$ denotes the number of components, D represents datasets, GT represents the ground truth number of classes for each dataset.}
\vspace{-5mm}
\label{tab:cluster}
\end{table*}

\subsection{Performance Evaluation of Supervised Semantic Hashing}
We evaluate the performance of supervised hashing in this section. Table \ref{tab:su_results} shows the performances of different supervised hashing models on three datasets under different lengths of hashing codes. We observe that all of the VAE-based generative hashing models (i.e VDSH, NASH, GMSH and BMSH) exhibit better performance, demonstrating the effectiveness of generative models on the task of semantic hashing. It can be also seen that BMSH-S achieves the best performance, suggesting that the advantages of Bernoulli mixture priors can also be extended to the supervised scenarios.

To gain a better understanding about the relative performance gain of the four proposed models, the retrieval precisions of GMSH, BMSH, GMSH-S and BMSH-S using 32-bit hashing codes on the three datasets are plotted together in Figure \ref{fig:supervised_ours}. It can be obviously seen that GMSH-S and BMSH-S outperform GMSH and BMSH by a substantial margin, respectively. This suggests that the proposed generative hashing models can also leverage the label information to improve the hashing codes' quality.

\subsection{ Impacts of the Component Number}\label{impact}
To investigate the impacts of component number, experiments are conducted for GMSH and BMSH under different values of $K$. For demonstration convenience, the length of hashing codes is fixed to 32. Table \ref{tab:cluster} shows the precisions of top 100 retrieved documents when the number of components $K$ is set to different values. We can see that the retrieval precisions of the proposed models, especially the BMSH, are quite robust to this parameter. For BMSH, the difference between the best and worst precisions on the three datasets are 0.0123, 0.0052 and 0.0134, respectively, which are small comparing to the gains that BMSH has achieved. One exception is the performance of GMSH on 20Newsgroups dataset. However, as seen from Table \ref{tab:cluster}, as long as the number $K$ is not too small, the performance loss is still acceptable. It is worth noting that the worst performance of GMSH on 20Newsgroups is 0.4708, which is still better than VDSH's 0.4327 as in Table \ref{tab:results}. For the BMSH model, the performance is stable across all the considered datasets and $K$ values.

\subsection{Visualization of Learned Embeddings}
To understand the performance gains of the proposed models better, we visualize the learned representations of VDSH-S, GMSH-S and BMSH-S on 20Newsgroups dataset. UMAP ~\cite{mcinnes2018umap-software} is used to project the 32-dimensional latent representations into a 2-dimensional space, as shown in Figure \ref{fig:embedding}. Each data point in the figure denotes a document, with each color representing one category. The number shown with the color is the ground truth category ID. It can be observed from Figure \ref{fig:embedding} (a) and (b) that more embeddings are clustered correctly when the Gaussian mixture prior is used. This confirms the advantages of using mixture priors in the task of hashing. Furthermore, it is observed that the latent embeddings learned by BMSH-S can be clustered almost perfectly. In contrast, many embeddings are found to be clustered incorrectly for the other two models. This observation is consistent with the conjecture that mixture prior and end-to-end training are both useful for semantic hashing.

\section{Conclusions}
In this paper, deep generative models with mixture priors were proposed for the tasks of semantic hashing. We first proposed to use a Gaussian mixture prior, instead of the standard Gaussian prior in VAE, to learn the representations of documents. A separate step was then used to cast the continuous latent representations into binary hashing codes. To avoid the requirement of a separate casting step, we further proposed to use the Bernoulli mixture prior, which offers the advantages of both mixture prior and the end-to-end training. Comparing to strong baselines on three public datasets, the experimental results indicate that the proposed methods using mixture priors outperform existing models by a substantial margin. Particularly, the semantic hashing model with Bernoulli mixture prior (BMSH) achieves state-of-the-art results on all the three datasets considered in this paper.

\section{Acknowledgements}
This work is supported by the National Natural Science Foundation of China (NSFC) under Grant No. 61806223, U1711262, U1501252, U1611264 and U1711261, and National Key R\&D Program of China (2018YFB1004404).

\nocite{fu2019cyclical}

\bibliography{emnlp}

\begin{thebibliography}{31}
\expandafter\ifx\csname natexlab\endcsname\relax\def\natexlab#1{#1}\fi

\bibitem[{Bengio et~al.(2013)Bengio, L{\'e}onard, and
  Courville}]{bengio2013estimating}
Yoshua Bengio, Nicholas L{\'e}onard, and Aaron Courville. 2013.
\newblock Estimating or propagating gradients through stochastic neurons for
  conditional computation.
\newblock \emph{arXiv preprint arXiv:1308.3432}.

\bibitem[{Bowman et~al.(2015)Bowman, Vilnis, Vinyals, Dai, Jozefowicz, and
  Bengio}]{bowman2015generating}
Samuel~R Bowman, Luke Vilnis, Oriol Vinyals, Andrew~M Dai, Rafal Jozefowicz,
  and Samy Bengio. 2015.
\newblock Generating sentences from a continuous space.
\newblock \emph{arXiv preprint arXiv:1511.06349}.

\bibitem[{Chaidaroon et~al.(2018)Chaidaroon, Ebesu, and
  Fang}]{chaidaroon2018deep}
Suthee Chaidaroon, Travis Ebesu, and Yi~Fang. 2018.
\newblock Deep semantic text hashing with weak supervision.
\newblock SIGIR.

\bibitem[{Chaidaroon and Fang(2017)}]{chaidaroon2017variational}
Suthee Chaidaroon and Yi~Fang. 2017.
\newblock Variational deep semantic hashing for text documents.
\newblock In \emph{Proceedings of the 40th International ACM SIGIR Conference
  on Research and Development in Information Retrieval}, pages 75--84. ACM.

\bibitem[{Chen et~al.(2016)Chen, Kingma, Salimans, Duan, Dhariwal, Schulman,
  Sutskever, and Abbeel}]{chen2016variational}
Xi~Chen, Diederik~P Kingma, Tim Salimans, Yan Duan, Prafulla Dhariwal, John
  Schulman, Ilya Sutskever, and Pieter Abbeel. 2016.
\newblock Variational lossy autoencoder.
\newblock \emph{arXiv preprint arXiv:1611.02731}.

\bibitem[{Datar et~al.(2004)Datar, Immorlica, Indyk, and
  Mirrokni}]{datar2004locality}
Mayur Datar, Nicole Immorlica, Piotr Indyk, and Vahab~S Mirrokni. 2004.
\newblock Locality-sensitive hashing scheme based on p-stable distributions.
\newblock In \emph{Proceedings of the twentieth annual symposium on
  Computational geometry}, pages 253--262. ACM.

\bibitem[{Dong et~al.(2004)Dong, Halevy, Madhavan, Nemes, and
  Zhang}]{dong2004similarity}
Xin Dong, Alon Halevy, Jayant Madhavan, Ema Nemes, and Jun Zhang. 2004.
\newblock Similarity search for web services.
\newblock In \emph{Proceedings of the Thirtieth international conference on
  Very large data bases-Volume 30}, pages 372--383. VLDB Endowment.

\bibitem[{Fu et~al.(2019)Fu, Li, Liu, Gao, Celikyilmaz, and
  Carin}]{fu2019cyclical}
Hao Fu, Chunyuan Li, Xiaodong Liu, Jianfeng Gao, Asli Celikyilmaz, and Lawrence
  Carin. 2019.
\newblock Cyclical annealing schedule: A simple approach to mitigating kl
  vanishing.
\newblock In \emph{Proceedings of the 2019 Conference of the North American
  Chapter of the Association for Computational Linguistics: Human Language
  Technologies, Volume 1 (Long and Short Papers)}, pages 240--250.

\bibitem[{Gong et~al.(2013)Gong, Lazebnik, Gordo, and
  Perronnin}]{gong2013iterative}
Yunchao Gong, Svetlana Lazebnik, Albert Gordo, and Florent Perronnin. 2013.
\newblock Iterative quantization: A procrustean approach to learning binary
  codes for large-scale image retrieval.
\newblock \emph{IEEE Transactions on Pattern Analysis and Machine
  Intelligence}, 35(12):2916--2929.

\bibitem[{Goyal et~al.(2017)Goyal, Hu, Liang, Wang, and
  Xing}]{goyal2017nonparametric}
Prasoon Goyal, Zhiting Hu, Xiaodan Liang, Chenyu Wang, and Eric~P Xing. 2017.
\newblock Nonparametric variational auto-encoders for hierarchical
  representation learning.
\newblock In \emph{Proceedings of the IEEE International Conference on Computer
  Vision}, pages 5094--5102.

\bibitem[{Jiang et~al.(2016)Jiang, Zheng, Tan, Tang, and
  Zhou}]{jiang2016variational}
Zhuxi Jiang, Yin Zheng, Huachun Tan, Bangsheng Tang, and Hanning Zhou. 2016.
\newblock Variational deep embedding: An unsupervised and generative approach
  to clustering.
\newblock \emph{arXiv preprint arXiv:1611.05148}.

\bibitem[{Kingma and Ba(2014)}]{kingma2014adam}
Diederik~P Kingma and Jimmy Ba. 2014.
\newblock Adam: A method for stochastic optimization.
\newblock \emph{arXiv preprint arXiv:1412.6980}.

\bibitem[{Kingma and Welling(2013)}]{kingma2013auto}
Diederik~P Kingma and Max Welling. 2013.
\newblock Auto-encoding variational bayes.
\newblock \emph{arXiv preprint arXiv:1312.6114}.

\bibitem[{Koren(2008)}]{koren2008factorization}
Yehuda Koren. 2008.
\newblock Factorization meets the neighborhood: a multifaceted collaborative
  filtering model.
\newblock In \emph{Proceedings of the 14th ACM SIGKDD international conference
  on Knowledge discovery and data mining}, pages 426--434. ACM.

\bibitem[{Kulis and Darrell(2009)}]{kulis2009learning}
Brian Kulis and Trevor Darrell. 2009.
\newblock Learning to hash with binary reconstructive embeddings.
\newblock In \emph{Advances in neural information processing systems}, pages
  1042--1050.

\bibitem[{Lew et~al.(2006)Lew, Sebe, Djeraba, and Jain}]{lew2006content}
Michael~S Lew, Nicu Sebe, Chabane Djeraba, and Ramesh Jain. 2006.
\newblock Content-based multimedia information retrieval: State of the art and
  challenges.
\newblock \emph{ACM Transactions on Multimedia Computing, Communications, and
  Applications (TOMM)}, 2(1):1--19.

\bibitem[{Liu et~al.(2016)Liu, Wang, Shan, and Chen}]{liu2016deep}
Haomiao Liu, Ruiping Wang, Shiguang Shan, and Xilin Chen. 2016.
\newblock Deep supervised hashing for fast image retrieval.
\newblock In \emph{Proceedings of the IEEE conference on computer vision and
  pattern recognition}, pages 2064--2072.

\bibitem[{Liu et~al.(2012)Liu, Wang, Ji, Jiang, and Chang}]{liu2012supervised}
Wei Liu, Jun Wang, Rongrong Ji, Yu-Gang Jiang, and Shih-Fu Chang. 2012.
\newblock Supervised hashing with kernels.
\newblock In \emph{Computer Vision and Pattern Recognition (CVPR), 2012 IEEE
  Conference on}, pages 2074--2081. IEEE.

\bibitem[{Liu et~al.(2011)Liu, Wang, Kumar, and Chang}]{liu2011hashing}
Wei Liu, Jun Wang, Sanjiv Kumar, and Shih-Fu Chang. 2011.
\newblock Hashing with graphs.
\newblock In \emph{Proceedings of the 28th international conference on machine
  learning (ICML-11)}, pages 1--8. Citeseer.

\bibitem[{McInnes et~al.(2018)McInnes, Healy, Saul, and
  Grossberger}]{mcinnes2018umap-software}
Leland McInnes, John Healy, Nathaniel Saul, and Lukas Grossberger. 2018.
\newblock Umap: Uniform manifold approximation and projection.
\newblock \emph{The Journal of Open Source Software}, 3(29):861.

\bibitem[{Miao et~al.(2016)Miao, Yu, and Blunsom}]{miao2016neural}
Yishu Miao, Lei Yu, and Phil Blunsom. 2016.
\newblock Neural variational inference for text processing.
\newblock In \emph{International Conference on Machine Learning}, pages
  1727--1736.

\bibitem[{Shen et~al.(2018)Shen, Su, Chapfuwa, Wang, Wang, Carin, and
  Henao}]{shen2018nash}
Dinghan Shen, Qinliang Su, Paidamoyo Chapfuwa, Wenlin Wang, Guoyin Wang,
  Lawrence Carin, and Ricardo Henao. 2018.
\newblock Nash: Toward end-to-end neural architecture for generative semantic
  hashing.
\newblock \emph{arXiv preprint arXiv:1805.05361}.

\bibitem[{Shen et~al.(2015)Shen, Shen, Liu, and Tao~Shen}]{shen2015supervised}
Fumin Shen, Chunhua Shen, Wei Liu, and Heng Tao~Shen. 2015.
\newblock Supervised discrete hashing.
\newblock In \emph{Proceedings of the IEEE conference on computer vision and
  pattern recognition}, pages 37--45.

\bibitem[{Stein et~al.(2007)Stein, zu~Eissen, and
  Potthast}]{stein2007strategies}
Benno Stein, Sven~Meyer zu~Eissen, and Martin Potthast. 2007.
\newblock Strategies for retrieving plagiarized documents.
\newblock In \emph{Proceedings of the 30th annual international ACM SIGIR
  conference on Research and development in information retrieval}, pages
  825--826. ACM.

\bibitem[{Wang et~al.(2018)Wang, Zhang, Sebe, Shen et~al.}]{wang2018survey}
Jingdong Wang, Ting Zhang, Nicu Sebe, Heng~Tao Shen, et~al. 2018.
\newblock A survey on learning to hash.
\newblock \emph{IEEE Transactions on Pattern Analysis and Machine
  Intelligence}, 40(4):769--790.

\bibitem[{Wang et~al.(2016)Wang, Liu, Kumar, and Chang}]{wang2016learning}
Jun Wang, Wei Liu, Sanjiv Kumar, and Shih-Fu Chang. 2016.
\newblock Learning to hash for indexing big data—a survey.
\newblock \emph{Proceedings of the IEEE}, 104(1):34--57.

\bibitem[{Wang et~al.(2013)Wang, Zhang, and Si}]{wang2013semantic}
Qifan Wang, Dan Zhang, and Luo Si. 2013.
\newblock Semantic hashing using tags and topic modeling.
\newblock In \emph{Proceedings of the 36th international ACM SIGIR conference
  on Research and development in information retrieval}, pages 213--222. ACM.

\bibitem[{Weiss et~al.(2009)Weiss, Torralba, and Fergus}]{weiss2009spectral}
Yair Weiss, Antonio Torralba, and Rob Fergus. 2009.
\newblock Spectral hashing.
\newblock In \emph{Advances in neural information processing systems}, pages
  1753--1760.

\bibitem[{Xu et~al.(2015)Xu, Wang, Tian, Xu, Zhao, Wang, and
  Hao}]{xu2015convolutional}
Jiaming Xu, Peng Wang, Guanhua Tian, Bo~Xu, Jun Zhao, Fangyuan Wang, and
  Hongwei Hao. 2015.
\newblock Convolutional neural networks for text hashing.
\newblock In \emph{IJCAI}, pages 1369--1375.

\bibitem[{Yang et~al.(2017)Yang, Hu, Salakhutdinov, and
  Berg-Kirkpatrick}]{yang2017improved}
Zichao Yang, Zhiting Hu, Ruslan Salakhutdinov, and Taylor Berg-Kirkpatrick.
  2017.
\newblock Improved variational autoencoders for text modeling using dilated
  convolutions.
\newblock \emph{arXiv preprint arXiv:1702.08139}.

\bibitem[{Zhang et~al.(2010)Zhang, Wang, Cai, and Lu}]{zhang2010self}
Dell Zhang, Jun Wang, Deng Cai, and Jinsong Lu. 2010.
\newblock Self-taught hashing for fast similarity search.
\newblock In \emph{Proceedings of the 33rd international ACM SIGIR conference
  on Research and development in information retrieval}, pages 18--25. ACM.

\end{thebibliography}
\bibliographystyle{acl_natbib}

\end{document}